\def\year{2015}
\documentclass[letterpaper]{article}
\usepackage{aaai}
\usepackage{times}
\usepackage{helvet}
\usepackage{courier}
\usepackage{algorithm}
\usepackage{algorithmicx}
\usepackage{amsmath}
\usepackage{amsthm}
\usepackage{graphicx}
\begin{document}
%
\title{Sparse Deep Stacking Network for Image Classification}
\author{Jun Li, Heyou Chang, Jian Yang\\
School of Computer Science and Technology \\
Nanjing University of Science and Technology, Nanjing, 219000, China.\\
junl.njust@gmail.com;changheyou891001@126.com;csjyang@njust.edu.cn\\
}
\maketitle
\begin{abstract}
\begin{quote}
Sparse coding can learn good robust representation to noise and model more higher-order representation for image classification. However, the inference algorithm is computationally expensive even though the supervised signals are used to learn compact and discriminative dictionaries in sparse coding techniques. Luckily, a simplified neural network module (SNNM) has been proposed to directly learn the discriminative dictionaries for avoiding the expensive inference. But the SNNM module ignores the sparse representations. Therefore, we propose a sparse SNNM module by adding the mixed-norm regularization ($l_1/l_2$ norm). The sparse SNNM modules are further stacked to build a sparse deep stacking network (S-DSN). In the experiments, we evaluate S-DSN with four databases, including Extended YaleB, AR, 15 scene and Caltech101. Experimental results show that our model outperforms related classification methods with only a linear classifier. It is worth noting that we reach $98.8\%$ recognition accuracy on 15 scene.
\end{quote}
\end{abstract}

\section{Introduction}
It is well-known that sparse representations have a number of theoretical and practical advantages in computer vision and machine learning \cite{Lee2007,Gregor2010,Yang2012}. In particular, sparse coding techniques have led to promising results in image classification, e.g. face recognition and digit classification. Sparse coding, as a generative model, is a very important way to extract the sparse representations. However, sparse coding has the expensive inference algorithm and does not use the label of the training data. Although some researchers use the supervised signals to learn compact and discriminative dictionaries \cite{Jiang2013,Zhuang2013,Huang2013}, the expensive inference algorithm is still a problem. Since it is to train the dictionaries by using the labels, do we directly learn the discriminative dictionaries for avoiding the expensive inference?

Fortunately, a simplified neural network module (SNNM) \cite{Deng2011a} can directly train the discriminative dictionaries and fast calculate the representations. In the SNNM, the input layer is non-linearly mapped to a hidden layer by using a projection matrix $\textbf{W}$ and a sigmoid activation function, and linearly mapped to an output layer by a matrix $\textbf{U}$. Clearly, $\textbf{W}$ has discriminative ability because it is trained by minimizing the least squares error between the output vector and label vector. Moreover, SNNM can fast infer the hidden representation by only calculating a projection multiplication and a nonlinear transformation. Following a stacked scheme \cite{Wolpert1992}, many SNNM modules are further stacked to build a Deep Stacking Network (DSN), which is previously named the Deep Convex Network \cite{Deng2011b}. Recently, DSN has received increasing attentions due to its successful application in speech classification and information retrieval \cite{Deng2012,Deng2013a}. Additionally, the DSN is attractive in that SNNM's the batch-mode nature offers a potential solution to the insurmountable problem of scalability in dealing with virtually unlimited amount of training data available nowadays \cite{Deng2013book}. Therefore, we extend DSN for image classification.

Despite DSN's success in speech classification, its framework also has several limitations. First, the conventional DSN only has used the sigmoid activation function for the nonlinear hidden layer \cite{Deng2012}. Although sigmoid has been widely used in the literature, it suffers from a number of drawbacks: for example the training can be slow, and with random initialization, the solution can stuck at a poor local solution that does not have good predictive performance \cite{Glorot2010}. In fact there are another two types of activation functions. The one is {\em hyperbolic tangent}, which has been applied to training deep neural networks. It suffers from the same problems as those of sigmoid functions. A more recent proposal is the {\em rectifier linear unit} (ReLU) \cite{Nair2010}. It is observed that this method is very useful for object recognition and often trains significantly faster \cite{Glorot2011}.

Second, sparse representations play a key role in image classification because they have the power to learn good robust features to noise, train gabor-like filters, and model more higher-order features \cite{Ranzato2007,Lee2008}. Evidently, sparse representations have led to promising results in image classification \cite{Jiang2013}. Furthermore, there is considerable evidence that in brain the percentage of neurons active is between 1 and 4\% \cite{Lennie2003}. It is reasonable to consider the sparse representations in SNNM modules. However, the conventional techniques for training SNNM completely ignores the sparse representations. Generally, they can be achieved by penalizing non-zero activation of hidden units \cite{Ranzato2008} or a deviation of the expected activation of the hidden units \cite{Lee2008} in neural networks.
Moreover, in neural networks the local dependencies between hidden units can make hidden units for better modeling observed data \cite{Luo2011}. But SNNM module restricted connections within hidden layer can not exhibit these dependencies. Fortunately, the hidden units without increasing the connections can be divided into non-overlapping groups for capturing the local dependencies among hidden units \cite{Luo2011}. The local dependencies can be implemented by using $l_1/l_2$ regularization upon the activation possibilities of hidden units in SNNM module.

In light of the above argument, this paper exploits a Sparse Deep Stacking Network (S-DSN) for image classification. S-DSN is obtained by stacking the sparse SNNM modules, which consider the two activation function: ReLU and sigmoid; and use the group sparse penalties ($l_1/l_2$ regularization) to penalize the hidden representations in SNNM modular. Our S-DSN has many explicit advantages. First, compared with sparse coding technique (LC-KSVD \cite{Jiang2013}), one-layer S-DSN can learn the projection dictionaries, which lead to a faster inference. Second, compared with DSN, S-DSN can extract sparse representations for learning good features in image classification. Last, S-DSN can retain the scalable structure of DSN. To conform the advantages of the S-DSN for image classification, extensive experiments have been performed on the four databases, including Extended YaleB, AR, 15 scene and Caltech101. Compared with multiple related methods, the experiments show that our model gets better classification results than other benchmark methods. In particular, we reach $98.8\%$ recognition accuracy on 15 scene.

\section{Deep Stacking Network}
\label{sec:dsn}
The DSN architecture is originally presented in the literature \cite{Deng2011b}. Deng and Yu explore an original strategy for building deep networks, based on stacking layers of the basic SNNM modules, which take the simplified form of multilayer perceptron. We mathematically describe as follow:

Let the target vectors $\textbf{t}_i=[t_{1i},\cdots,t_{ji},\cdots,t_{Ci}]^T$ be arranged to form the columns of $\textbf{T}=[\textbf{t}_1,\cdots,\textbf{t}_i,\cdots,\textbf{t}_N]\in R^{C\times N}$. Let the input data vectors $\textbf{x}_i=[x_{1i},\cdots,x_{ji},\cdots,x_{Di}]^T$ be arranged to form the columns of $\textbf{X}=[\textbf{x}_1,\cdots,\textbf{x}_i,\cdots,\textbf{x}_N]\in R^{D\times N}$. Formally, in the basic module, the lower-layer weight matrix, which is denoted by $\textbf{W}\in R^{D\times L}$, connects the linear input layer and the nonlinear hidden layer. The upper-layer weight matrix, which is denoted by $\textbf{U}\in R^{L\times C}$, connects the nonlinear hidden layer with the linear output layer. The outputs of upper-layer is $\textbf{Y}=\textbf{U}^T\textbf{H}$, where $\textbf{H}=\sigma(\textbf{W}^T\textbf{X})\in R^{L\times N}$ is the hidden layer outputs and $\sigma(a)=1/(1 + e^{-a})$ is the sigmoid activation function \cite{Deng2011a,Deng2011b}. The parameters $\textbf{U}$ and $\textbf{W}$ are learned to minimize the least squares objective:
\begin{equation}
\min_{\textbf{U},\textbf{W}}f_{dsn} = \|\textbf{U}^T\textbf{H}-\textbf{T}\|^2_F+\alpha\|\textbf{U}\|^2_F
\label{eq:dsn}
\end{equation}
where $\alpha$ is a regularization parameter of upper-layer weight matrix $\textbf{U}$.

Clearly, $\textbf{U}$ has a closed-form solution:
\begin{equation}
\textbf{U}=(\textbf{H}\textbf{H}^T+\alpha \textbf{I})^{-1}\textbf{H}\textbf{T}^T
\label{eq:dsnsloveu}
\end{equation}

By using a gradient descent \cite{Deng2011b} algorithm to minimize the the least squares objective in \eqref{eq:dsn} and deriving the gradient of $\textbf{W}$ in the basic module, we obtain
\begin{equation}
\frac{\partial f_{dsn} }{\partial \textbf{W}}=2\textbf{X}\left[\textbf{H}^T\circ(\textbf{I}-\textbf{H}^T)\circ(\textbf{U}\textbf{U}^T\textbf{H}-\textbf{U}\textbf{T})^T\right]
\label{eq:dsnslovew}
\end{equation}
where $\circ$ denotes element-wise multiplication and $\textbf{I}$ is the matrix of all ones.

The "convex" solution accentuates the role of convex optimization in learning the output network weights $\textbf{U}$ in each basic module \cite{Deng2011a}. Many basic modules are often stacked up with one on top of another to form a deep model. More specifically, the input units of a higher module can include the output units of the lowest module and optionally the raw input feature in the DSN \cite{Deng2011b}. For obtaining the higher-order information in the data, DSN has recently been extended to Tensor-DSN (T-DSN) \cite{Hutchinson2013}, which's the basic module is to replace a linear map from hidden representation to output with a bilinear mapping. It retains the scalable structure of DSN and provides the higher-order feature interactions missing in DSN.

\section{Sparse Deep Stacking Network}
\label{sec:sdsn}
The S-DSN is a sparse case of the DSN. The stacking operation of the S-DSN is the same as that for the DSN described in \cite{Deng2011b}. The general paradigm is to use the output vector of lower module and the original input vector to form the expanded "input" vector for the higher module of the DSN. The modular architecture of S-DSN is different from that of DSN. We consider the sigmoid function and the ReLU function; and the sparse penalties are added into the hidden units of modular architecture.

\subsection{Sparse Module}
\label{sec:sparsemodule}
The output of upper-layer is $\textbf{Y}=\textbf{U}^T\widehat{\textbf{H}}$ and the hidden layer output is as follow:
\begin{equation}
\widehat{\textbf{H}}=\phi(\textbf{W}^T\textbf{X})\in R^{L\times N}
\label{eq:hiddenrep}
\end{equation}
where $\phi(a)$ is the sigmoid activation function $\sigma(a)$ or the ReLU activation function $\max(0,a)$. For simplicity, let $\mathcal{H}=\{1,2,\cdots,L\}$ denote the set of all hidden units. $\mathcal{H}$ is are divided into $G$ groups, where $G$ is the number of groups. The $g$th group is denoted by $\mathcal{G}_g$, where $\mathcal{H}=\bigcup_{g=1}^G\mathcal{G}_g$ and $\bigcap_{g=1}^G\mathcal{G}_g=\emptyset$. So,
$\widehat{\textbf{H}}=[\widehat{\textbf{H}}_{\mathcal{G}_1,:};\cdots;\widehat{\textbf{H}}_{\mathcal{G}_g,:};\cdots;\widehat{\textbf{H}}_{\mathcal{G}_G,:}]$.

The parameters $\textbf{U}$ and $\textbf{W}$ are learned to minimize the least squares objective:
\begin{equation}
\min_{\textbf{U},\textbf{W}}f_{sdsn} = \|\textbf{U}^T\widehat{\textbf{H}}-\textbf{T}\|^2_F+\alpha\|\textbf{U}\|^2_F + \beta \Psi(\widehat{\textbf{H}})
\label{eq:sdsn}
\end{equation}
where $\alpha$ is a regularization parameter of upper-layer weight matrix $\textbf{U}$, $\beta$ is a regularization constant of the activation of the hidden units and $\Psi(\widehat{\textbf{H}})$ represents the imposed penalty over sparse representations $\widehat{\textbf{H}}$. Typically, the $l_1$ norm is conducted as a penalty to explicitly enforce sparsity on each sparse representation. It is described as:
\begin{equation}
\Psi(\widehat{\textbf{H}})=\sum_{i=1}^N \|\widehat{\textbf{h}}_{i}\|_1
\label{eq:sparsel1}
\end{equation}
where $\widehat{\textbf{h}}_{i}$ is the representation of $i$th example $(i=1,\cdots,N)$.

In neural networks, sparse representations are advantageous for classification \cite{Ranzato2007}. Moreover, group sparse representations \cite{SBengio2009} can learn the statistical dependencies between hidden units and lead to better performance \cite{Luo2011}. To implement the dependencies, we averagely divide hidden units into non-overlapping groups to restrain the dependencies within these groups and force hidden units in a group to compete with each other \cite{Luo2011}. Luckily, a mixed-norm regularization ($l_1/l_2$-norm) can be conducted in the modular architecture to achieve group sparse representations. Following \cite{Luo2011}, we consider the mixed-norm regularization, which is as follows:
\begin{equation}
\Psi(\widehat{\textbf{H}})= \sum_{g=1}^G\|\widehat{\textbf{H}}_{\mathcal{G}_g,:}\|_{1,2}
\label{eq:sparsel1l2}
\end{equation}
where $\widehat{\textbf{H}}_{\mathcal{G}_g,:}$ is the representation matrix associated to those intra-modality data belonging to the $g$th group and the $l_1/l_2$-norm is defined as
\begin{equation}
\|\widehat{\textbf{H}}_{\mathcal{G}_g,:}\|_{1,2}=\sum_{i=1}^N\sqrt{\sum_{j\in\mathcal{G}_g}\widehat{h}_{j,i}^2}
\end{equation}

\subsection{Learning Weights- Algorithm}

Once the lower-layer weight matrix $\textbf{W}$ are fixed, $\widehat{\textbf{H}}$ are also determined uniquely. Then solving the upper-layer weight matrix $\textbf{U}$ can be formulated as a convex optimization problem:
\begin{equation}
\min_{\textbf{U}}f^u_{sdsn} = \|\textbf{U}^T\widehat{\textbf{H}}-\textbf{T}\|^2_F+\alpha\|\textbf{U}\|^2_F
\label{eq:sdsnu}
\end{equation}
which has a closed-form solution:
\begin{equation}
\textbf{U}=(\widehat{\textbf{H}}\widehat{\textbf{H}}^T+\alpha \textbf{I})^{-1}\widehat{\textbf{H}}\textbf{T}^T
\label{eq:sdsnsloveu}
\end{equation}

There are two algorithms for learning the lower-layer weight matrix $\textbf{W}$. First, given fixed current $\textbf{U}$, $\textbf{W}$ can be optimized using a gradient descent algorithm \cite{Deng2011a} to minimize the squared error objective:
\begin{equation}
\min_{\textbf{W}}f^1_{sdsn} = \|\textbf{U}^T\widehat{\textbf{H}}-\textbf{T}\|^2_F+\beta \Psi(\widehat{\textbf{H}})
\label{eq:sdsnw1}
\end{equation}
and deriving the gradient, we obtain
\begin{align}
\frac{\partial f^1_{sdsn} }{\partial \textbf{W}}=&2\textbf{X}\left[d\phi(\widehat{\textbf{H}}^T)\circ(\textbf{U}\textbf{U}^T\widehat{\textbf{H}}-\textbf{U}\textbf{T})^T\right] \nonumber \\
& + 2\beta\textbf{X}\left[d\phi(\widehat{\textbf{H}}^T)\circ\widehat{\textbf{H}}^T\circ/\widetilde{\textbf{H}}^T\right]
\label{eq:sdsnslovew1}
\end{align}
where $\circ$ denotes element-wise multiplication, $\circ/$ denotes element-wise division, $\widetilde{\textbf{H}}$ that it's element is $\widetilde{h}_{j,i}=\sqrt{\sum_{j\in\mathcal{G}_g}\widehat{h}_{j,i}^2}$, $d\phi(\widehat{\textbf{H}}^T)$ denotes element-wise gradient computation, $d\phi(a)$ is the gradient of the activation function. When $\phi(a)$ is the sigmoid activation function, $d\phi(a)=\sigma(a)\times(1-\sigma(a))$ and when $\phi(a)$ is the ReLU activation function, $d\phi(a)$ is described as:
\begin{equation}
 d\phi(a)= \left\{
   \begin{array}{ll}
   1, & \hbox{$a>0$;} \\
   0, & \hbox{$a\leq 0$.}
   \end{array}
   \right.
\label{eq:dphi}
\end{equation}
To ReLU activation function, we follow the hypothesis \cite{Glorot2011} that the hard non-linearities do not hurt the optimization so long as the gradient can be propagated to many hidden units.

\begin{algorithm}[tb]
\caption{Training Algorithm of Sparse Modular}
\begin{algorithmic}[1]
\State {\bfseries Input:} Data matrix $\textbf{X}$, label matrix $\textbf{T}$, parameters $\theta=\{\epsilon, \alpha, \beta, G \}$ and training epochs $E$.
\State {\bfseries Initialize:} Projection Matrix $\textbf{W}$ are initialized with small random values.
\State  Given $\textbf{W}$, calculate $\widehat{\textbf{H}}$ by Eq. \eqref{eq:hiddenrep}.
\State  Update $\textbf{W}$ by Eq. \eqref{eq:sdsnupdataw}.
\State  Repeat 3-4 $E$ epochs (or until convergence).
\State {\bfseries Output} weight matrix $\textbf{W}$.
\end{algorithmic}
\label{alg:Trainalgsm}
\vskip -0.05in
\end{algorithm}

Second, for faster moving $\textbf{W}$ towards a direction that finds the optimal points, the deterministic nonlinear relationship between $\textbf{U}$ and $\textbf{W}$ is used to compute the gradient. By plugging \eqref{eq:sdsnsloveu} into criterion \eqref{eq:sdsn}, the least squares objective is rewritten as:
\begin{align}
\min_{\textbf{W}}f^2_{sdsn} = &\|[(\widehat{\textbf{H}}\widehat{\textbf{H}}^T+\alpha \textbf{I})^{-1}\widehat{\textbf{H}}\textbf{T}^T]^T\widehat{\textbf{H}}-\textbf{T}\|^2_F+ \nonumber\\
                             & \alpha\|(\widehat{\textbf{H}}\widehat{\textbf{H}}^T+\alpha \textbf{I})^{-1}\widehat{\textbf{H}}\textbf{T}^T\|^2_F +\beta \Psi(\widehat{\textbf{H}})
\label{eq:sdsnw2}
\end{align}

However, when regularization is used in the objective function \eqref{eq:sdsn} (i.e. $\alpha>0$), the gradient of $f^2_{sdsn}$ can be very complicated.
To simplify the gradient we assume $\alpha=0$ in $f^2_{sdsn}$. So, second term of $f^2_{sdsn}$ is equivalent to zero. Similar to \cite{Deng2011b}, then we derive the gradient $\frac{\partial f^2_{sdsn}}{\partial \textbf{W}}$ and obtain
\begin{align}
\frac{\partial f^2_{sdsn}}{\partial \textbf{W}}= & 2\textbf{X}\left[d\phi(\widehat{\textbf{H}}^T)\circ[\widehat{\textbf{H}}^\dag(\widehat{\textbf{H}}\textbf{T}^T)(\textbf{T}\widehat{\textbf{H}}^\dag)-
\textbf{T}^T(\textbf{T}\widehat{\textbf{H}}^\dag)]\right] \nonumber\\
&\ \ \ + 2\beta\textbf{X}\left[d\phi(\widehat{\textbf{H}}^T)\circ\widehat{\textbf{H}}^T\circ/\widetilde{\textbf{H}}^T\right]
\label{eq:sdsnslovew2}
\end{align}
where $\widehat{\textbf{H}}^\dag=\widehat{\textbf{H}}^T(\widehat{\textbf{H}}\widehat{\textbf{H}}^T)^{-1}$ and $d\phi(\cdot)$ and $\widetilde{\textbf{H}}$ are defined in \eqref{eq:sdsnslovew1}.

The algorithm then updates $\textbf{W}$ using the gradient defined in \eqref{eq:sdsnslovew1} and \eqref{eq:sdsnslovew2} as
\begin{align}
\textbf{W}=\textbf{W}-\epsilon \frac{\partial f^1_{sdsn} }{\partial \textbf{W}} \ \
or \ \ \textbf{W}=\textbf{W}-\epsilon \frac{\partial f^2_{sdsn} }{\partial \textbf{W}}
\label{eq:sdsnupdataw}
\end{align}
where $\epsilon$ is a learning rate. The weight matrix learning process is outlined in \textbf{Algorithm 1}. 

\begin{algorithm}[tb]
\caption{Training Algorithm of S-DSN}
\begin{algorithmic}[1]
\State {\bfseries Input:} Data $\textbf{X}$, label $\textbf{T}$, parameters $\theta=\{\epsilon, \alpha, \beta, G\}$, training epochs $E$ and the number of layers $K$.
\State {\bfseries Initialize:} $\textbf{X}^1=\textbf{X}$ and $k=1$.
\State {\bfseries while} $k\leq K$
\State  Given $\textbf{X}^k$, $\textbf{T}$, $\theta$ and $E$, optimize $\textbf{W}^k$ by \textbf{Algorithm 1}.
\State  Given $\textbf{X}^k$ and $\textbf{W}^k$, calculate $\widehat{\textbf{H}}^k$ by Eq. \eqref{eq:hiddenrep}, $\textbf{U}^k$ by Eq. \eqref{eq:sdsnsloveu} and $\textbf{Y}^k=\left(\textbf{U}^k\right)^T\widehat{\textbf{H}}^k$.
\State $\textbf{X}^{k+1}=[\textbf{X}; \textbf{Y}^k]$.
\State {\bfseries end while}
\State {\bfseries Output} weight matrix $\textbf{W}^k(k=1,\cdots,K)$.
\end{algorithmic}
\label{alg:TrainalgS-DSN}
\vskip -0.05in
\end{algorithm}

\subsection{The S-DSN Architecture}
The spare SNNM module described in the above subsection is used to construct the $K$-layers S-DSN architecture, where $K$ is the number of layers. In $k$th spare module we denote the input by $\textbf{X}^k$, hidden representations by $\widehat{\textbf{H}}^k$, output by $\textbf{Y}^k$, label matrix by $\textbf{T}$ and weight matrix by $\textbf{W}^k$ and $\textbf{U}^k$. Given input data $\textbf{X}$ and label $\textbf{T}$, when $k=1$, $\textbf{X}^1=\textbf{X}$. Then the general paradigm of S-DSN can be decomposed in three phases:
\begin{itemize}
  \item Step 1: Train the $k$th sparse module to minimize the least squares error between $\textbf{Y}^k$ and $\textbf{T}$.
  \item Step 2: Generate the input $\textbf{X}^{k+1}$ of the $k+1$th sparse module by adding the output $\textbf{Y}^k$ of $k$th sparse module.
  \item Step 3: Iterate as in Step 1 and Step 2 to construct the S-DSN architecture.
\end{itemize}

We summarize the optimization of S-DSN in \textbf{Algorithm 2}.
For capturing the spare representation from raw data, this paper proposes the S-DSN, which is implemented by penalizing the hidden unit activations and rectifying the negative of outputs of hidden units activations. Due to the simple structure of each module, the S-DSN still retains the computational advantage of the DSN in parallelism and scalability during learning all parameters.

\section{Experiments}
\label{sec:experiment}
We present experimental results on four databases: the Extended YaleB database, the AR face database, Caltech101 and 15 scene categories.
\begin{itemize}
  \item Extended YaleB database: this database contains 2,414 frontal face images of 38 people. There are about 64 images for each person. The original images were cropped and normalized to $192\times 168$ pixels.
  \item AR database: this database consists of over 4,000 color images of 126 people. Each person has 26 face images taken during two sessions. These images include more facial variations, including different illumination conditions, different expressions, and different facial "disguises" (sunglasses and scarves). Following the standard evaluation procedure, we use a subset of the database consisting of 2,600 images from 50 male subjects and 50 female subjects. Each face image was also cropped and normalized to $165\times 120$ pixels.
  \item Caltech-101: this database [10] contains 9144 images belonging to 101 classes, with about 40 to 800 images per class. Most images of Caltech-101 are with medium resolution, i.e., about $300\times300$.
  \item 15-Scene: this data set, compiled by several researchers [11, 20, 24], contains a total of 4485 images falling into 15 categories, with the number of images per category ranging from 200 to 400. The categories include living room, bedroom, kitchen, highway, mountain and et al.
\end{itemize}

According to \cite{Jiang2013}, the four databases are preprocessed \footnote{they can be downloaded from: http://www.umiacs.umd.edu/~zhuolin/projectlcksvd.html}: in the Extended YaleB database and AR face database, each face image is projected onto a $n$-dimensional feature vector with a randomly generated matrix from a zero-mean normal distribution. The dimension of a random-face feature in Extended YaleB is 504 while the dimension in AR face is 540. In face databases the $n$-dimensional features of each image are normalized to $[-1,1]$. For the Caltech101 database, we first extract sift descriptors from $16\times16$ patches, which are densely sampled from each image on a dense grid with 6-pixels stepsize; then we extract the spatial pyramid feature based on the extracted sift features with three grids of size $1\times1$, $2\times2$ and $4\times4$. To train the codebook for spatial pyramid, we use the standard $k$-means clustering with $k = 1024$. For the 15 scene category database, we compute the spatial pyramid feature using a four-level spatial pyramid and a SIFT-descriptor codebook with a size of 200. Finally, the spatial pyramid features are reduced to 3000 dimensions by PCA.

The matrix parameters are initialized with small random values sampled from a normal distribution with zero mean and standard deviation of $0.01$. For simplicity, we use the constant learning rate $\epsilon$ chosen from $\{20, 15, 5, 2, 1, 0.2, 0.1, 0.05, 0.01, 0.001\}$, the regularization parameter $\alpha$ chosen from $\{1, 0.5, 0.1\}$, the sparse regularization parameter $\beta$ chosen from $\{0.1, 0.05, 0.01, 0.001, 0.0001\}$ and the group number $G$ chosen from $\{2, 4, 5, 10, 20\}$. In all experiments, we only train 5 epochs, the number of hidden units is 500 and the number of layers is 2. For each data set, each experiment is repeated 10 times with random selected training and testing images, and the average precision is reported.
In the rest of this paper, we denote that S-DSN(sigm) indicates S-DSN with sigmoid function; S-DSN(relu) indicates S-DSN with ReLU function; DSN-1, S-DSN(sigm)-1 and S-DSN(relu)-1 respectively indicate one-layer DSN, S-DSN(sigm) and S-DSN(relu); DSN-2, S-DSN(sigm)-2 and S-DSN(relu)-2 respectively indicate two-layer DSN, S-DSN(sigm) and S-DSN(relu).

\subsection{Sparseness Comparisons}
\label{sec:sparsecom}

\begin{table}[t]
\small{\caption{Hoyer's sparseness measures (HSM) on Extended YaleB and AR databases. We train on 15 (10) samples per category for Extended YaleB (AR) and the rest for testing. For two databases, the number of hidden units is 500, the group sizes for S-DSN(sigm) and S-DSN(relu) are 4 and the number of layers is 2. In Extended YaleB, $\epsilon=0.2$ and $\alpha=0.5$ are used for DSN; $\epsilon=0.2$, $\alpha=0.5$ and $\beta=0.001$ are used for S-DSN(sigm). In Extended YaleB $\epsilon=0.05$ and $\alpha=1$ are used for DSN; $\epsilon=0.05$, $\alpha=1$ and $\beta=0.0001$ are used for S-DSN(relu).}}
\label{tab:hspm}
\begin{center}
\small{
\begin{tabular}{c|c|c|c|c|c}
\hline
\multicolumn{2}{c|}{} & \multicolumn{2}{c|}{S-DSN(sigm)} & \multicolumn{2}{c}{DSN} \\
\hline
\multicolumn{1}{c|}{}         & layers   & HSM &Acc. ($\%$) & HSM &Acc. ($\%$)   \\
\multicolumn{1}{c|}{Extended} & 1  &0.096& 91.4  &0.010& 88.9    \\
\multicolumn{1}{c|}{YaleB}    & 2  &0.111& 92.0  &0.012& 89.4  \\
\hline
\multicolumn{2}{c|}{} & \multicolumn{2}{c|}{S-DSN(relu)} & \multicolumn{2}{c}{DSN} \\
\hline
\multicolumn{1}{c|}{}        & layers   & HSM &Acc. ($\%$) & HSM &Acc. ($\%$) \\
\multicolumn{1}{c|}{AR}      & 1  &0.286& 93.2  &0.003& 80.2  \\
\multicolumn{1}{c|}{ }       & 2  &0.306& 93.5  &0.003& 81.2  \\
\hline
\end{tabular}}
\end{center}
\vskip -0.2in
\end{table}

Before presenting classification results, we first show the sparseness of S-DSN(sigm) and S-DSN(relu) compared to DSN. We use Hoyer's sparseness measure (HSM) \cite{Hoyer2004} to figure out how sparse representations learned by the S-DSN(sigm), S-DSN(relu) and DSN. This measure has good properties, which is in the interval $[0, 1]$ and on a normalized scale. Its value more close to $1$ means that there are more zero components in the vector. We perform comparisons on Extended YaleB and AR databases, and results are reported in Table 1. The sparseness results show that S-DSN(sigm) and S-DSN(relu) have higher sparseness and higher recognition accuracy. Table 1 compares the network HSM of the S-DSN(sigm) and the S-DSN(relu) to that of DSN. We observe that the average sparseness of two layers S-DSN(sigm) is about 0.105 (Extended YaleB) and the average sparseness of two layers S-DSN(relu) is about 0.291 (AR). In contract, the average sparseness of two layers DSN is on average below $0.02$ in the databases. It can be seen that the S-DSN can learn sparser representations. Due to space reasons, Figure 1 only visualizes the activation probabilities of first hidden layer, which are computed under the S-DSN(relu) and the DSN given an image from test set of AR.

\begin{figure}[ht]
\vskip -0.1in
\centering
\centerline{\includegraphics[width=0.82\columnwidth]{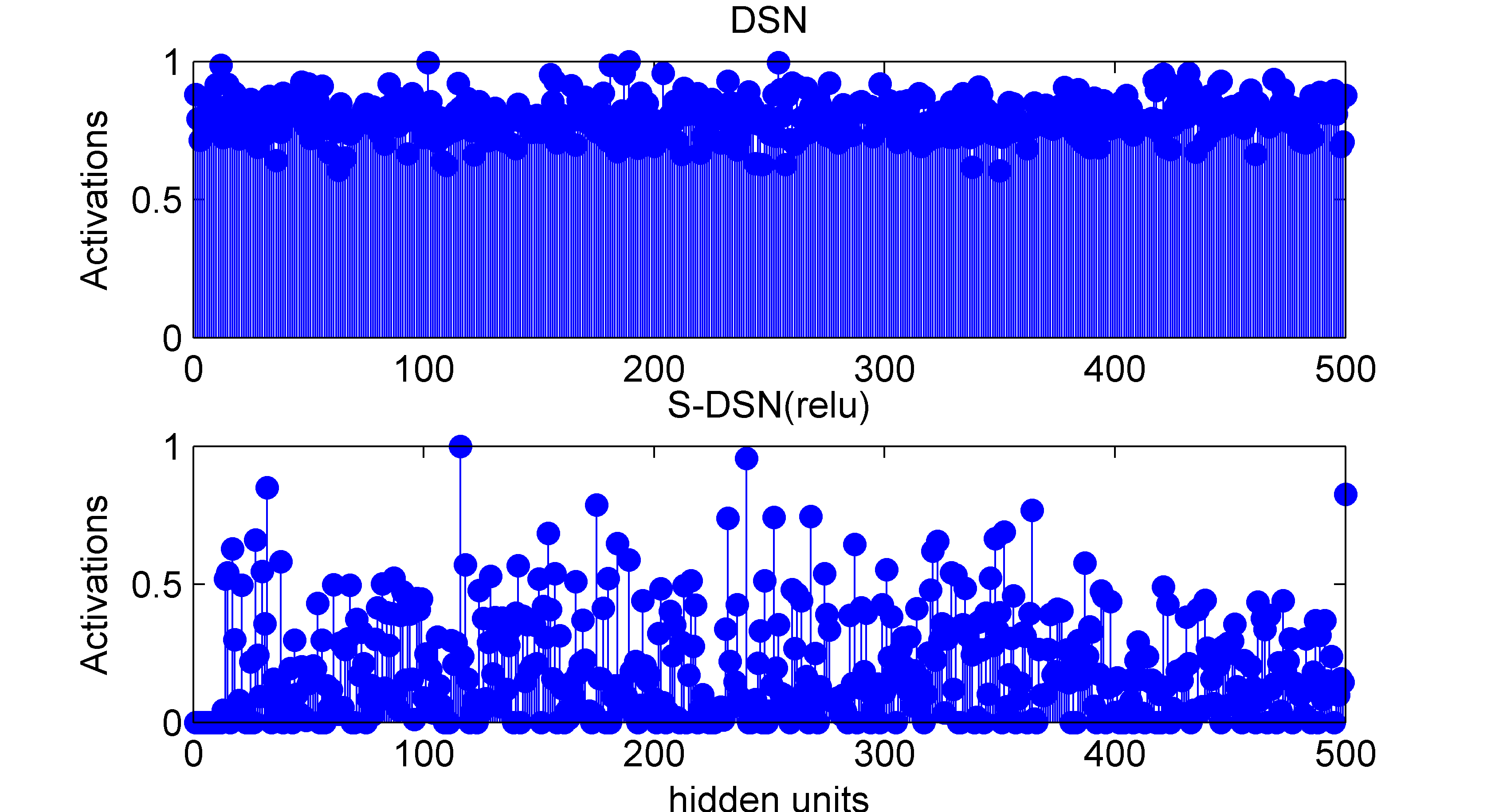}}
\vskip -0.1in
\small{
\caption{Activation probabilities of first hidden layer are computed under DSN and S-DSN(relu) on the AR database. Activation probabilities be normalized by dividing the maximum of activation probabilities.}}
\label{fig:hsmar}
\vskip -0.15in
\end{figure}

\subsection{Results}
\label{sec:result}

\begin{table}[t]
\small{\caption{Recognition Results Using Random-Face Features on the Extended YaleB Database}}
\label{tab:resultExtYaleB}
\begin{center}
\small{
\begin{tabular}{c|c|c|c}
\hline
Methods      &    Acc. ($\%$) &  Methods       &  Acc. ($\%$)  \\
\hline
SRC          &   97.2             &  LC-KSVD       &   96.7     \\
DSN-1        &   96.6             &  DSN-2         &   96.9     \\
\hline
S-DSN(sigm)-1     &  96.9              & S-DSN(sigm)-2       &  \textbf{97.4}   \\
S-DSN(relu)-1     &  96.1              & S-DSN(relu)-2       &  96.7    \\
\hline
\end{tabular}}
\end{center}
\vskip -0.25in
\end{table}

\begin{table}[t]
\small{\caption{Recognition Results Using Random Face Features on the AR Face Database}}
\label{tab:resultAR}
\begin{center}
\small{
\begin{tabular}{c|c|c|c}
\hline
Methods   &Acc. ($\%$) &  Methods      &  Acc. ($\%$)     \\
\hline
SRC       &   97.5         & LC-KSVD       &   97.8       \\
DSN-1    &   97.6         & DSN-2        &   97.8       \\
\hline
S-DSN(sigm)-1  &  97.9          & S-DSN(sigm)-2      &  \textbf{98.1}  \\
S-DSN(relu)-1  &  97.6          & S-DSN(relu)-2      &  97.8           \\
\hline
\end{tabular}}
\end{center}
\vskip -0.25in
\end{table}

\begin{table}
\vskip -0.2in
\small{\caption{Inference Time (ms) for a Test Image on the Extended YaleB Database}}
\label{tab:inftimeExtYaleB}
\begin{center}
\small{
\begin{tabular}{c|c|c|c}
\hline
  Methods         & SRC  &    LC-KSVD  &   S-DSN(relu)    \\
\hline
Average time  & 20.121 &  0.502       &     0.069  \\
\hline
\end{tabular}}
\end{center}
\end{table}

\textbf{Face Recognition}
Extended YaleB: We randomly select half (32) of the images per category as training and the other half for testing. The parameters are selected as follow: in DSN $\epsilon=0.1$ and $\alpha=0.5$; in S-DSN(sigm) $\epsilon=0.1$, $\alpha=0.5$, $G=2$, and $\beta=0.01$; in S-DSN(relu) $\epsilon=0.01$, $\alpha=2$, $G=5$, and $\beta=0.001$.
AR: For each person, we randomly select 20 images for training and the other 6 for testing. In our experiments, $\epsilon=0.1$ and $\alpha=0.5$ are used in DSN; $\epsilon=0.1$, $\alpha=0.5$, $G=4$, and $\beta=0.001$ are used in S-DSN(sigm); $\epsilon=0.01$, $\alpha=1$, $G=4$, and $\beta=0.001$ are used in S-DSN(relu).

We compare S-DSN with DSN \cite{Deng2011b}, and LC-KSVD \cite{Jiang2013} and SRC \cite{Wright2009} algorithms, which reported state-of-the-art results on those two databases. The experimental results are summarized in Table 2 and Table 3, respectively. S-DSN(sigm) achieves better results than DSN, LC-KSVD and SRC. From Table 2 S-DSN(sigm)-1 is better than LC-KSVD and has about $0.2\%$ improvement in Extended YaleB. From Table 3, S-DSN(sigm)-1 and S-DSN(sigm)-2 are also better than LC-KSVD and have about $0.1\%$ and $0.3\%$ improvement in AR, respectively. In addition, we compare with LC-KSVD in terms of the computation time for classifying one test image.  S-DSN has a faster inference because it can directly learn projection dictionaries. As shown in Table 4, S-DSN is 7 times faster than LC-KSVD.

\textbf{15 Scene Category:}
Following the common experimental settings, we randomly choose 100 images from each class for training data and the rest for test data. In our experiments, $\epsilon=20$ and $\alpha=0.1$ are used in DSN; $\epsilon=20$, $\alpha=0.1$, $G=4$, and $\beta=0.05$ are used in S-DSN(sigm); $\epsilon=15$, $\alpha=0.1$, $G=4$, and $\beta=0.001$ are used in S-DSN(relu).

\begin{table}[t]
\small{\caption{Recognition Results Using Spatial Pyramid Features on the 15 Scene Category Database}}
\label{tab:resultsc15}
\begin{center}
\small{
\begin{tabular}{c|c|c|c}
\hline
Methods       &    Acc. ($\%$) &  Methods        &  Acc. ($\%$)  \\
\hline
ITDL          &   81.1             & ISPR+IFV        &   91.1    \\
SR-LSR        &   85.7             & ScSPM           &   80.3    \\
LLC           &   89.2             & SRC             &   91.8    \\
LC-KSVD       &   92.9             & DeepSC          &   83.8    \\
DeCAF         &   88.0             & DSFL+DeCAF      &   92.8    \\
DSN-1         &   96.7             & DSN-2           &   97.0    \\
\hline
S-DSN(sigm)-1      &   96.5             & S-DSN(sigm)-2        &  97.1    \\
S-DSN(relu)-1      &   \textbf{98.8}    & S-DSN(relu)-2        &  \textbf{98.8}    \\
\hline
\end{tabular}}
\end{center}
\vskip -0.2in
\end{table}

\begin{figure}[ht]
\begin{center}
\centerline{\includegraphics[width=0.87\columnwidth]{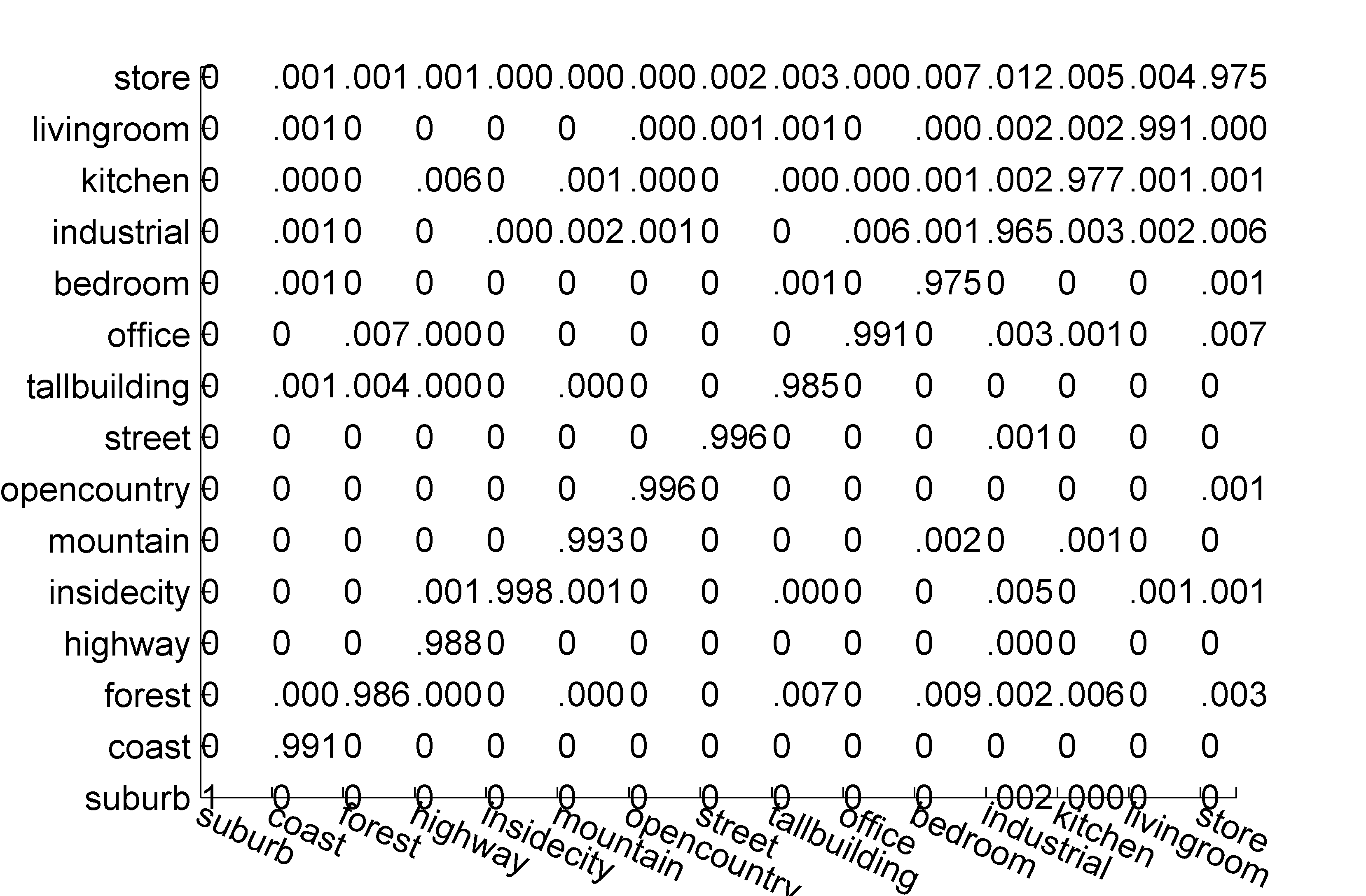}}
\vskip -0.15in
\small{\caption{The confusion matrix on the 15 scene category database.}}
\label{fig:cmsc15}
\end{center}
\vskip -0.35in
\end{figure}

We compare our results with SRC \cite{Wright2009}, LC-KSVD \cite{Jiang2013}, DeepSC \cite{He2014}, DSN \cite{Deng2011b} and other state-of-the-art approaches: ScSPM \cite{Yangjc2009}, LLC \cite{Wang2010}, ITDL \cite{Qiu2014}, ISPR+IFV \cite{Lin2014}, SR-LSR \cite{Lix2014}, DeCAF \cite{Donahue2014}, DSFL+DeCAF \cite{Zuo2014}. The detailed comparison results are shown in Table 5. Compared to LC-KSVD, S-DSN(relu)-1's performance is much better, since it makes a $5.9\%$ improvement. It also registers about $1.8\%$ improvement over the deep models: DeepSC, DeCAF, DSFL+DeCAF and DSN. As Table 5 shows, we see that S-DSN(relu) performs best among all existing methods. The confusion matrix for the S-DSN(relu) are further shown in Figure 2, from which we can see that the misclassification errors of industrial and store are higher than others.

\textbf{Caltech101:}
Following the common experimental settings, we train on 5, 10, 15, 20, 25, and 30 samples per category and test on the rest. Due to space reasons, we only give the parameters for 30 training samples per category: $\epsilon=0.2$ and $\alpha=0.5$ are used in DSN; $\epsilon=0.2$, $\alpha=0.5$, $G=4$, and $\beta=0.01$ are used in S-DSN(sigm); $\epsilon=0.05$, $\alpha=0.5$, $G=2$, and $\beta=0.001$ are used in S-DSN(relu).

We evaluate our approach using spatial pyramid features and compare with with SRC \cite{Wright2009}, LC-KSVD \cite{Jiang2013}, DeepSC \cite{He2014}, DSN \cite{Deng2011b} and other approaches ScSPM \cite{Yangjc2009}, LLC \cite{Wang2010}, LRSC \cite{Zhangt2013}, LCLR \cite{Jiang2014}. The average classification rates are reported in Table 6. From these results, S-DSN(relu)-1 outperforms the other competing dictionary learning approaches, including LC-KSVD, LRSC, and SRC; and has $1.6\%$ improvement. S-DSN(relu) also registers about $1.5\%$ improvement over a deep model: DSN. Note that $76.5\%$ accuracy achieved by our method (the number of hidden units is 1000) is also competitive with the $78.4\%$ reported in DeepSC.

\begin{table}[t]
\small{\caption{Recognition Results Using Spatial Pyramid Features on the Caltech101 Database}}
\label{tab:resultCaltech101}
\begin{center}
\small{
\begin{tabular}{c|c|c|c|c|c|c}
\hline
Methods        &    5        &     10       &      15      &     20       &     25      &    30         \\
\hline
ScSPM         &    -        &     -        &      67.0    &     -        &     -       &    73.2       \\
SRC            &    48.8     &    60.1      &      64.9    &     67.7     &     69.2    &    70.7       \\
LLC            &    51.2     &    59.8      &      65.4    &     67.7     &     70.2    &    73.4       \\
LC-KSVD        &    54.0     &    63.1      &      67.7    &     70.5     &     72.3    &    73.6       \\
LRSC           &    55.0     &    63.5      &      67.1    &     70.3     &     72.7    &    74.4       \\
LCLR           &    53.4     &    62.8      &      67.2    &     70.8     &     72.9    &    74.7       \\
DSN-1          &    53.6     &    61.8      &      67.7    &     70.2     &     72.0    &    74.6       \\
DSN-2          &    54.9     &    63.4      &      68.2    &     70.5     &     72.9    &    74.7       \\
\hline
S-DSN(sigm)-1       &    54.0     &    62.3      &      67.6    &     70.2     &     72.1    &    74.7       \\
S-DSN(sigm)-2       &    55.4     &    63.7      &      68.3    &     70.8     &     73.2    &    74.9       \\
S-DSN(relu)-1       &    55.4     &    63.8      &      68.7    &     71.2     &     73.5    &    76.0       \\
S-DSN(relu)-2       &\textbf{55.6}&\textbf{64.2} &\textbf{69.0} &\textbf{71.3} &\textbf{73.6}& \textbf{76.2} \\
\hline
\end{tabular}}
\end{center}
\vskip -0.2in
\end{table}

\begin{figure}[ht]
\begin{center}
\centerline{\includegraphics[width=0.7\columnwidth]{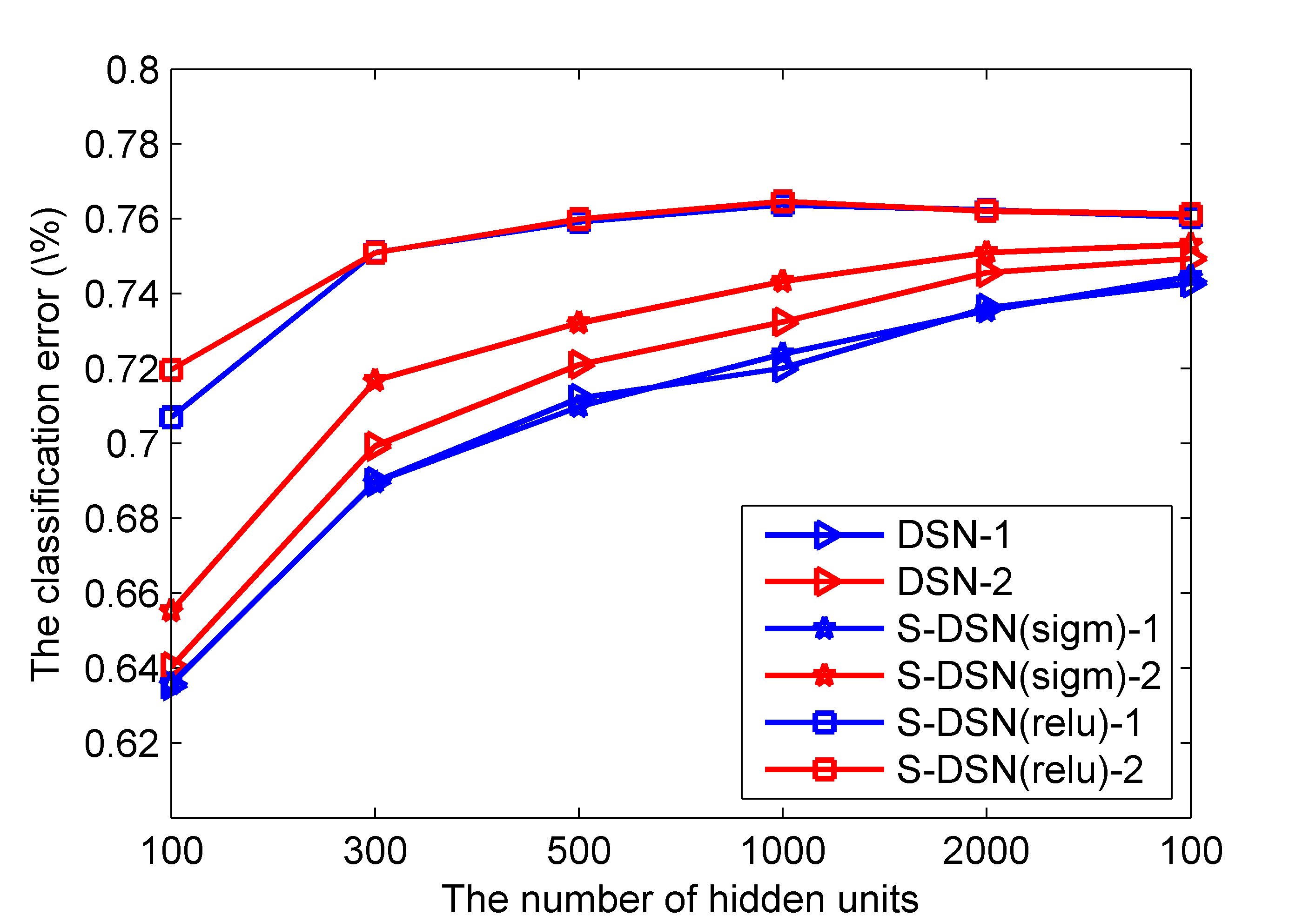}}
\vskip -0.15in
\small{\caption{Effect of the number of hidden units used in S-DSN(sigm), S-DSN(relu) and DSN on recognition accuracy. }}
\label{fig:dhidnumc101}
\end{center}
\vskip -0.35in
\end{figure}

We examine how performance of the proposed S-DSN changes when varying the number of hidden units. We randomly select 30 images per category for training data and the rest for test data. We consider six settings where the number of hidden units changes from 100 to 3000 and compare the results with DSN. As reported the results in Figure 3, our approaches maintain high classification accuracies and outperform the DSN model. When increasing the number of hidden units, the accuracy of the system improves for S-DSN(sigm), S-DSN(relu) and DSN.

\textbf{Effects of Number of Layers:} The deep framework utilizes multiple-layers of feature abstraction to get a better representation for images. From Tables  2, 3, 5 and 6, we check the effect of varying the number of layers and the classification accuracy improves as the number of layers increases. In addition, compared to the dictionary learning approaches, S-DSN has a faster inference and a deep architecture. Moreover, S-DSN has a good performance in image classification.

\section{Conclusion}
\label{sec:cons}

In this paper, we present an improved DSN model, S-DSN, for image classification. S-DSN is constructed by stacking many sparse SNNM modules. In each sparse SNNM module, the lower-layer weights and the upper-layer weights are solved by using the convex optimization and the gradient descent algorithm. We use the S-DSN to further extract the sparse representations from the random face features and spatial pyramid features for image classification. Experimental results show that S-DSN yields very good classification results on four public databases with only a linear classifier.

\section{Acknowledgments}
This work was partially supported by the National Science Fund for Distinguished Young Scholars under Grant Nos. 61125305, 61472187, 61233011 and 61373063, the Key Project of Chinese Ministry of Education under Grant No. 313030, the 973 Program No. 2014CB349303, Fundamental Research Funds for the Central Universities No. 30920140121005, and Program for Changjiang Scholars and Innovative Research Team in University No. IRT13072.

{\small
\bibliographystyle{aaai}
\bibliography{curRefs}
}

\end{document}